\documentclass{article}

\usepackage{arxiv} 

\usepackage[utf8]{inputenc} 
\usepackage[T1]{fontenc}    
\usepackage{hyperref}       
\usepackage{url}            
\usepackage{booktabs}       
\usepackage{amsfonts}       
\usepackage{nicefrac}       
\usepackage{microtype}      
\usepackage{lipsum}
\usepackage{tikz}
\usepackage{multirow}
\usepackage{amsmath}
\usepackage{graphics}
\usepackage{subcaption}
\usepackage{makecell}
\usepackage{hyperref}
\usepackage{xcolor}
\usepackage{rotating}
\usepackage{pgfplots}
\pgfplotsset{compat=1.8}
\usepackage{eurosym}      
\usepackage{filecontents} 
\usepackage{tabu}
\usepackage{tikz}
\usepackage{rotating}
\usepackage{graphicx}
\usepackage{gensymb}
\def\checkmark{\tikz\fill[scale=0.4](0,.35) -- (.25,0) -- (1,.7) -- (.25,.15) -- cycle;} 

\renewcommand{\headeright}{}
\date{}

\title{TORNADO-Net: mulTiview tOtal vaRiatioN semAntic segmentation with Diamond inceptiOn module}

\author{
  Martin Gerdzhev*, Ryan Razani, Ehsan Taghavi, Bingbing Liu\\
  Noah's Ark Lab, Huawei Technologies\\
  Canada, Toronto\\
  \texttt{ \{martin.gerdzhev, ryan.razani, ehsan.taghavi, liu.bingbing\}@huawei.com} \\
}

\begin{document}
\maketitle


\begin{abstract}
    Semantic segmentation of point clouds is a key component of scene understanding for robotics and autonomous driving. In this paper, we introduce TORNADO-Net - a neural network for 3D LiDAR point cloud semantic segmentation. We incorporate a multi-view (bird-eye and range) projection feature extraction with an encoder-decoder ResNet architecture with a novel diamond context block. Current projection-based methods do not take into account that neighboring points usually belong to the same class. To better utilize this local neighbourhood information and reduce noisy predictions, we introduce a combination of Total Variation, Lov\'asz-Softmax, and Weighted Cross-Entropy losses. We also take advantage of the fact that the LiDAR data encompasses $360 \degree$ field of view and uses circular padding. We demonstrate state-of-the-art results on the SemanticKITTI dataset and also provide thorough quantitative evaluations and ablation results. 
\end{abstract}






\section{Introduction}
    	

Semantic segmentation of point clouds, specially on data collected from LiDARs is becoming a crucial task in many applications such as robotics, autonomous driving, etc.  Semantic segmentation is a key component of a larger scope of tasks called scene understanding. The role of scene understanding is even more pronounced for autonomous systems such as self-driving cars, where safety is paramount and errors in the perception stack can lead to bad planning and accidents.

While semantic segmentation has been used on images in other domains, such as medical imaging and surveillance systems, it has seen limited applications in self-driving due to the difficulty of data annotation of the different sensors. An autonomous vehicle is equipped with many primary sensors, including cameras, LiDARs, RADARs and possibly other secondary sensors such as sonars, which help the task of perception for an autonomous system. The earliest works on data labeling in robotics and autonomous driving have been done on cameras and images due to their wide availability and somewhat easy labeling procedure. 

To enable $3$D data collected from LiDARs to be fully utilized in perception modules, one needs a labeled point cloud at the point level (semantic) to be available. To this date, SemanticKITTI \cite{DBLP:conf/iccv/BehleyGMQBSG19} proved to be one of the best LiDAR datasets with point wise labels available to researchers and industries. Due to this reason, the focus of this work is mainly on literature available on LiDAR semantic segmentation using SemanticKITTI \cite{DBLP:conf/iccv/BehleyGMQBSG19} as a benchmark dataset. It is worth noting that, due to wider field of view, precise distance measurements and light-invariance, most autonomous car platforms use LiDAR sensors for scene understanding on the road.

LiDAR point clouds present a number of challenges as compared to images - they are unstructured, sparse and their density varies with distance. Some methods try to tackle the problem by operating on the point clouds directly \cite{qi2017pointnet}, while others try to utilize the approaches from the image domain, by projecting the point clouds onto images (Bird Eye View, or Frontal projection) or 3D voxels, and applying convolutions on the structured representation \cite{wu2018squeezeseg}.

This work builds upon several approaches like multiple views projections similar to MVF \cite{zhou2020end} and encoder-decoder networks like SalsaNext \cite{cortinhal2020salsanext} to propose an end-to-end model that achieves state of the art results on SemanticKITTI \cite{DBLP:conf/iccv/BehleyGMQBSG19}. With TORNADO-Net we introduce the following contributions:
\begin{itemize}
  \item A pillar based learning module that learns and extracts features on BEV data representation of LiDAR point clouds;
  \item A novel global context module, named Diamond feature extractor, that processes the spherical range-image LiDAR data and extracts rich features suitable for semantic segmentation;
  \item A novel loss function based on total variation denoising techniques that improves the overall accuracy of point cloud semantic segmentation models;
  \item Circular padding to account for the LiDAR data with $360\degree$ horizontal field of view;
  \item An analysis on the semantic segmentation performance using different architectures through an extensive ablation study;
\end{itemize}

The rest of the paper is organized as follows. In Section \ref{sec:relwork}, a brief review of recent and related works is given. Section \ref{sec:citations} describes the proposed neural network model and the new loss functions in detail. Training details and experimental results including qualitative, quantitative, and ablation studies can be found in Section \ref{sec:result}, along with the benchmarks from all available and published methods in the literature for reference. Finally, conclusions and future work are presented in Section \ref{sec:conclusion}.
\section{Related Work}
\label{sec:relwork}


Until recently, there have been few approaches that focus on LiDAR point cloud based semantic segmentation due to the lack of large-scale labeled datasets. Some of these early methods include  PointNet \cite{qi2017pointnet}, SqueezeSeg \cite{wu2018squeezeseg} and DeepTemporalSeg \cite{dewan2019deeptemporalseg}.
The introduction of the SemanticKITTI dataset \cite{DBLP:conf/iccv/BehleyGMQBSG19} has spurred the development of novel LiDAR-based segmentation methods \cite{thomas2019kpconv, landrieu2018large, wu2019squeezesegv2, dewan2019deeptemporalseg, aksoy2019salsanet, rosu2019latticenet, milioto2019rangenet++, cortinhal2020salsanext, zhang2020polarnet, alonso20203d}. 

Based on how the point cloud is represented, methods can be grouped into methods that operate on points directly (point-wise), or methods that project the point cloud into a different, easier to work with structure (projection-based).
Point-wise methods like PointNet \cite{qi2017pointnet}, KPConv \cite{thomas2019kpconv} and RandLA-Net \cite{hu2019randla} don't require any preprocessing or transformation. Although capable of processing smaller point clouds, they are less useful for large point clouds (specially those which collect  $360\degree$ data) due to large memory requirements and slow inference speeds. To address some of these problems, some methods like SPG \cite{landrieu2018large} utilize a superpoint graph, which is formed by geometrically consistent elements. Some of the more prevailing methods, however, aim to project the pointcloud into either a 3D voxel grid, or a 2D image (Bird-Eye-View (BEV) or spherical Range View (RV)), and utilize convolutional operators that work on structured data \cite{wu2018squeezeseg, wu2019squeezesegv2, dewan2019deeptemporalseg, aksoy2019salsanet, rosu2019latticenet, milioto2019rangenet++, cortinhal2020salsanext, zhou2019endtoend}. The projection-based methods usually have achieved higher accuracy, while maintaining a much faster inference time. Therefore, TORNADONet aims at solving the problem of LiDAR semantic segmentation  using projection-based techniques that combines multiple projections -  first to BEV, and then to RV similar to \cite{zhou2019endtoend} to extract complementary features and achieve state-of-the-art results. 

In addition to the neural network model that is designed to address a specific problem, the choice of a loss function can play a crucial role to the accuracy of the model. For semantic segmentation tasks, cross entropy (CE) loss is one of the most widely used loss functions. However, since there can be a big class imbalance with many classes being over-represented in the data, the loss functions that take the class frequency into account, such as weighted cross entropy (WCE) or Focal loss \cite{lin2017focal}, can improve the performance. While these losses work well in a variety of tasks, they do not optimize the same criterion that is used to evaluate the performance of semantic segmentation, i.e., intersection-over-union or Jaccard Index (IoU). 

Since IoU is a discrete function and cannot be optimized for directly, surrogate functions that are differentiable have been proposed. Lov\'asz-Softmax loss \cite{berman2018lovasz} and Dice loss \cite{triess2020scanbased}, optimize the Jaccard and Dice coefficients, respectively in order to maximize the mean IoU(mIoU).

These losses however do not take into account the local neighbourhoods of the pixels/points and can lead to noisy predictions. To address this problem, regularization functions, such as, the Total Variation regularizer, have been used to regularize the overall loss by the smoothness of the prediction over neighboring pixels or data points. More recently, the TV regularizer was used to address the denoising problem in image applications \cite{chan2005recent, chen2010adaptive}. A major target in image denoising is to preserve important image features, such as edges, while removing noise. In our work, we use this concept and propose a novel TV loss function that takes into account neighbouring information.

To achieve better results, it is also common to use a weighted combination of multiple loss functions. For example, SalsaNext \cite{aksoy2019salsanet} uses a combination of WCE loss and Lov\'asz-Softmax loss. In this paper, a combination of WCE loss,  Lov\'asz-Softmax loss and the novel TV loss is proposed to help TORNADO-Net achieve state-of-the-art accuracy in LiDAR semantic segmentation.

\section{TORNADONet}
\label{sec:citations}
 \subsection{Model architecture}
 
\begin{figure*}[!htbp]
    \centering
    \includegraphics[trim = 0in 0in 0in 0in, clip, width=5.45in]{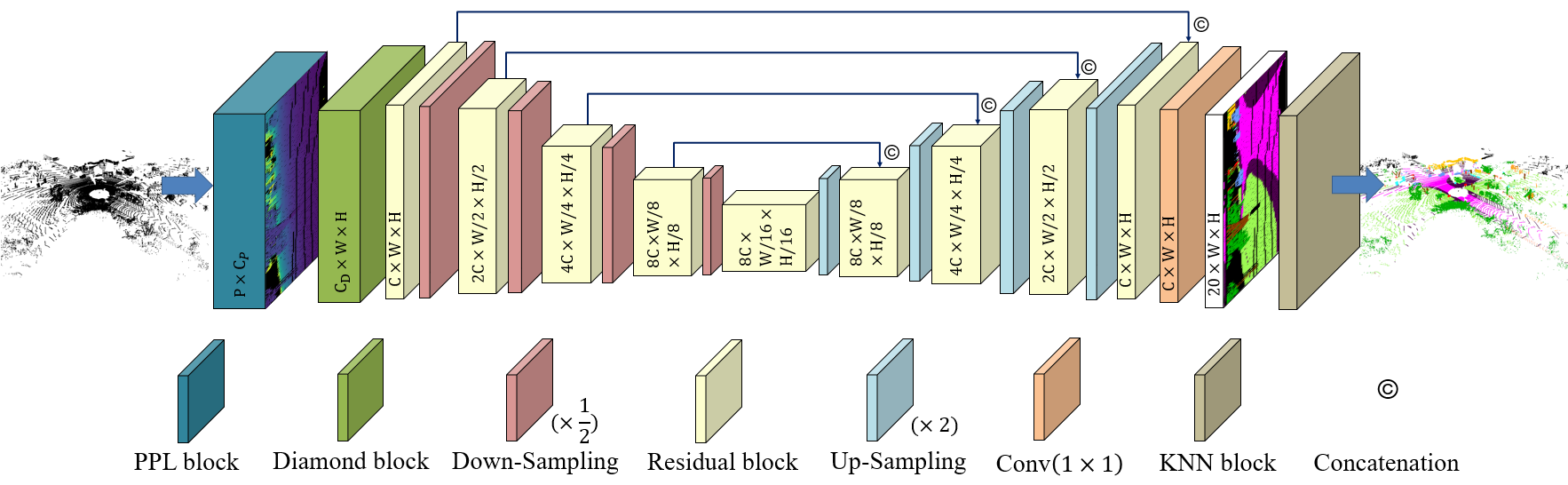}
    \caption[Tornado]{TORNADO Architecture.}
    \label{MVF}
\end{figure*}

 In this paper, a novel and intuitive NN architecture is introduced to solve the problem of LiDAR semantic segmentation benefiting from information extraction in different views. Although the encoder-decoder model processes range image similar to \cite{cortinhal2020salsanext}, the pillar-projection-learning module (PPL) learns and extracts information in the Bird's Eye View (BEV). This series of different projections as shown in Fig. \ref{MVF}, help the model pick up features that are otherwise difficult to extract. Results and ablation studies on the SemanticKITTI \cite{DBLP:conf/iccv/BehleyGMQBSG19} dataset benchmark show the effectiveness of the proposed CNN model.
 The details of the new architecture are explained in the subsections below.
 
 \subsubsection{Pillar Projection Learning}
\begin{figure*}[!htbp]
    \centering
    \includegraphics[trim = 0in 0in 0in 0in, clip, width=4in]{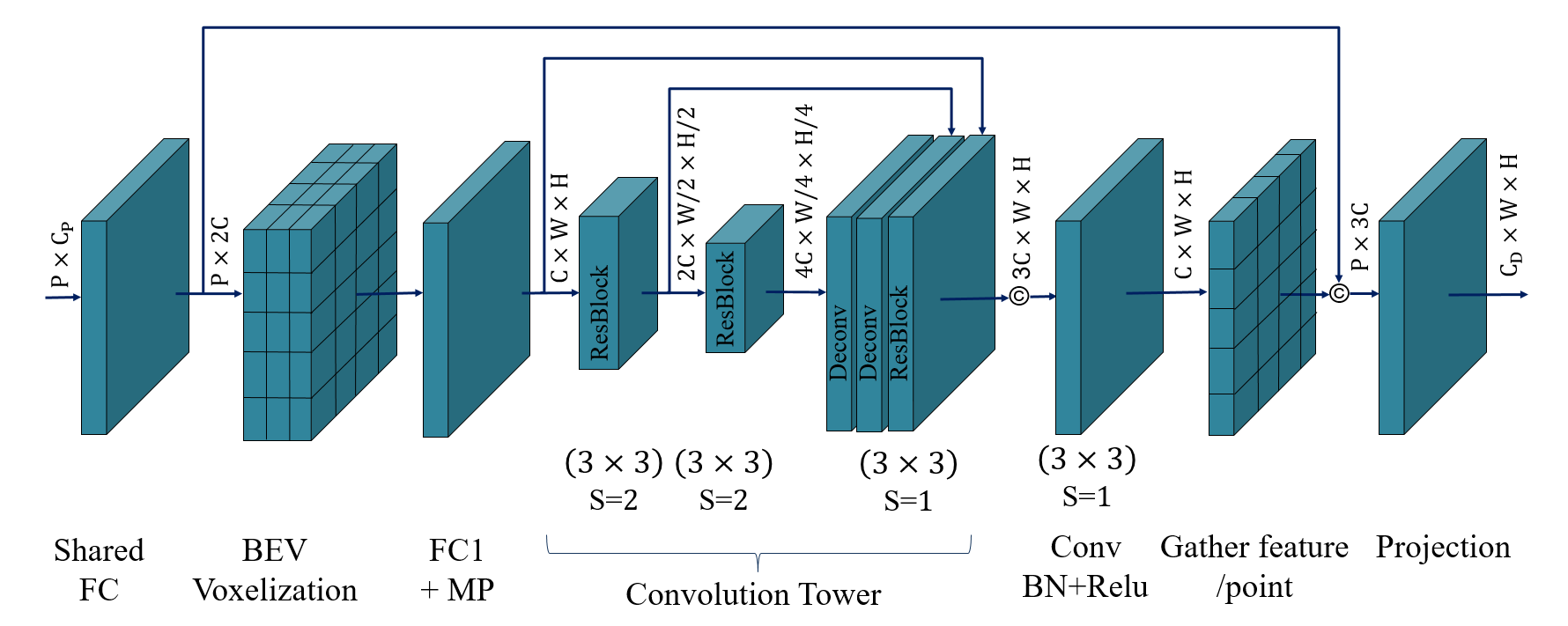}
    \caption[PPL]{PPL Block}
    \label{PPL}
\end{figure*}

 The proposed model starts off with the PPL module (see Figure \ref{PPL}) in which the raw LiDAR point cloud $P$ is processed by mapping the unordered points to BEV. We follow the approach similar to \cite{zhou2019endtoend} where points are grouped into pillars and their normalized $x,y,z$ pillar coordinates are appended to the point cloud features. The point cloud goes through a FC layer to extract better features and is then mapped to the BEV pillars. All points within a pillar are processed through another FC layer and then after a pooling operation each pillar is represented by a single feature vector. The projected pillars then undergo a series of strided convolutions followed by upsampling layers in order to capture neighbouring information. All points are then augmented with the features from the pillar that they belong to.
 
  The PPL thus generates rich point features that can be used for different applications. For the purpose of LiDAR semantic segmentation, range images have shown better and more accurate results with reasonable computational complexity. Because of this reason, in the proposed model and after applying PPL, the features are projected onto the range-image for further processing.
 
To accomplish this, one can use a transformation from point cloud to image as follows for $\mathbb{R}^3 \rightarrow \mathbb{R}^2$

    \begin{equation}
    \label{proj}
 	\left(
 	\begin{matrix}
 	u \\
 	v
 	\end{matrix}
 	\right) =
 	\left(
 	\begin{matrix}
 	\frac{1}{2}[1-arctan(y,x)\pi^{-1}]W \\
 	[1-(arcsin(zr^{-1})+f_{up})\frac{1}{f}]H
 	\end{matrix}
 	\right)
 \end{equation}
 where $(u,v)$ are the coordinates of a given point cloud $(x,y,z)$ in spherically transformed data, hereafter range-image, $W$ is the desired horizontal resolution and $H$ is the desired vertical resolution. Moreover, the vertical field of view of the sensor can be described as $f = f_{up} + f_{down}$. In its simplest form, the channels of the new range-image can be filled with $(x,y,z, rem, r)$, where $rem$ is remission reading of points and $r$ is the range. If desired, other channels can be added to the range-image to address the much needed features for a specific task. As the point cloud is already processed using a PPL block, in this NN model, we use extracted features $C_{D}$ and project them onto the range-image. Empty pixels in the transformed data can be masked.
 
 \subsubsection{Diamond contextual block}
 \label{sec:diamond}
\begin{figure*}[h]
    \centering
    \includegraphics[trim = 0in 0in 0in 0in, clip, width=3.5in]{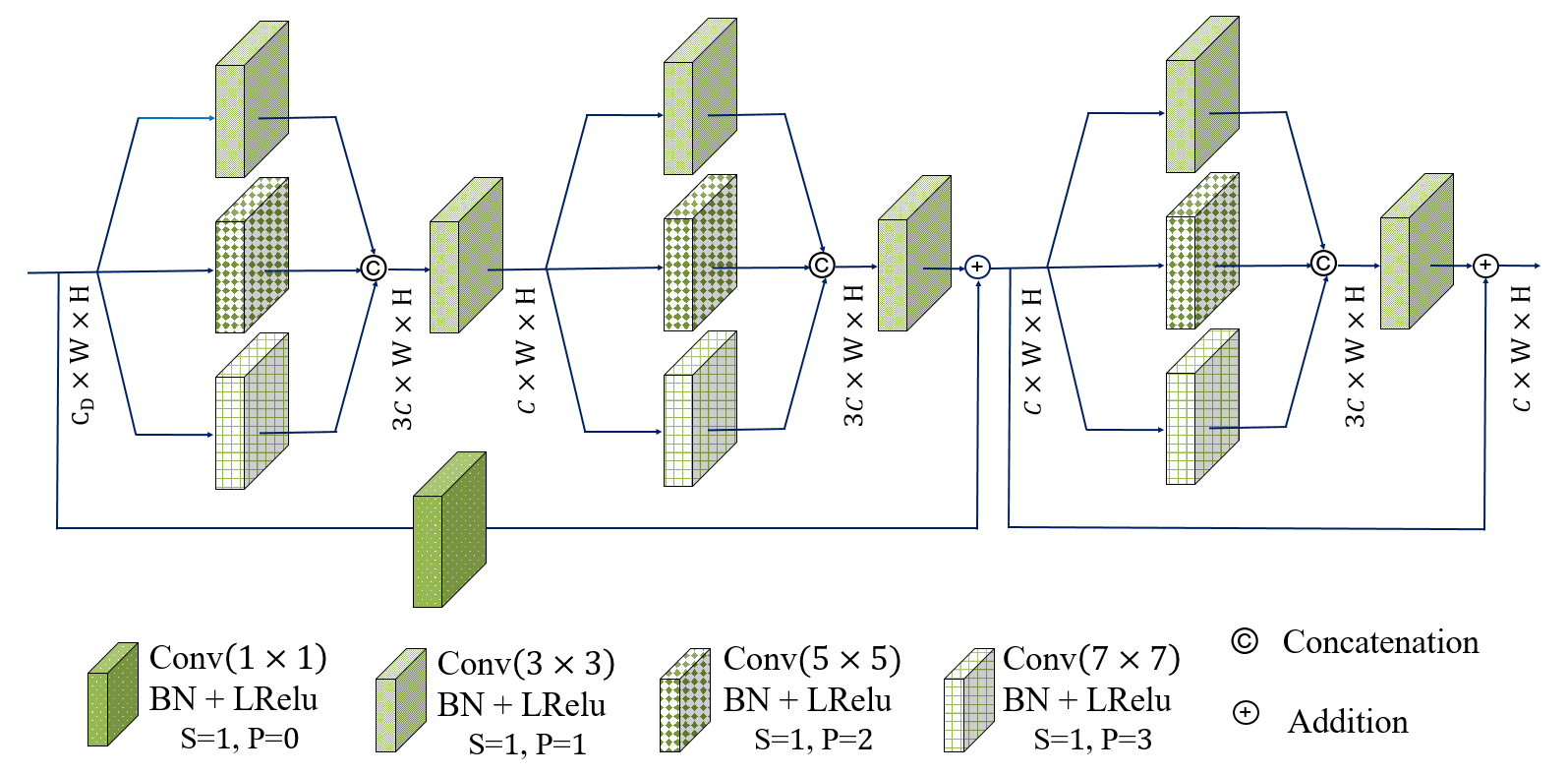}
    \caption[Diamond]{Diamond contextual block. This block processes the range-image output from the PPL block and provides rich contextual features which are suitable for LiDAR semantic segmentation.}
    \label{Diamond}
\end{figure*}

 Here we propose a new global context module to process the output of PPL, named diamond context block (DCB). DCB uses regular 2$D$ convolutions and provides context at different scales with efficient computation that can be used in various CNN models without loss. A generic block diagram of DCB is shown in Figure \ref{Diamond}. DCB consists of three diamond shape convolution blocks. Each block is a combination of $3\times 3$, $5\times 5$ and $7 \times 7$ $2$D convolutions. Moreover, to carry the local features, a skip connection with $1\times 1$ $2$D convolution connects the input to the output of the second diamond block. Finally, the third diamond block is concatenated with the skip connection of its input to generate the final feature tensor for further processing. Our ablation studies show that DCB enhances semantic segmentation on SemanticKITTI.

\subsubsection{Encoder-Decoder}
 After the raw point cloud is processed by PPL and DCB, the feature tensor is fed to an encoder-decoder CNN similar to what is proposed in \cite{aksoy2019salsanet}. The architecture of the proposed encoder-decoder is illustrated in Figure~\ref{MVF}. The input to the network is the spherical projection of the extracted features from PPL+DCB in section \ref{sec:diamond}. The encoder-decoder part of TORNADONet is built upon the base SalsaNet model \cite{aksoy2019salsanet} which follows the standard encoder-decoder architecture with a bottleneck compression rate of $16$. As opposed to the original implementation of SalsaNet with series of ResNet blocks \cite{he2016deep}, we use the blocks introduced in \cite{cortinhal2020salsanext} with dilations in the convolutions both on the decoder and encoder parts of the network. 

 \subsection{Loss function}\label{sub:loss}
One of the major contributions of this paper comes in the design of a new loss function using ideas introduced in \cite{zhang2018generalized}, \cite{berman2018lovasz} and \cite{rudin1992nonlinear}. In this section, we first briefly review  the common loss functions used in semantic segmentation tasks. Then the proposed loss function is introduced.


The weighted cross entropy loss \cite{zhang2018generalized, panchapagesan2016multi} can be written as,

\vspace{-2mm}
\begin{align}
{L}_{wce}(y,\hat{y})= - \sum_{i} a_{i} P(y_{i}) \log P(\hat{y}_{i}),\quad  a_{i}= 1/\sqrt(\nu_{i})
\end{align}
where $\nu_i$ is the frequency of each class, and $P(\hat{y}_{i})$ and $P(y_{i})$ are the corresponding predicted and ground truth probability. Note that  $P(y_{i})$ acts as indicator for the correct label class.

This loss is suitable where a NN model deals with multiclass classification problem much like semantic segmentation. Moreover, because it minimizes the distance between the two probability distributions, namely, predicted and actual, WCE makes sure that the difference between $P(\hat{y}_{i})$ and $P(y_{i})$ is being minimized.

The Lov\'asz-Softmax loss \cite{berman2018lovasz} can be expressed as:

\vspace{-1mm}
\begin{equation}
            \label{lovasz}
         	L_{ls} = \frac{1}{|C|}\sum_{c \in C}{J(\boldsymbol{e(c)})}
\end{equation}

\noindent where $J$ is the Lov\'asz extension of IoU, $\boldsymbol{e}(c)$ is the vector of errors for class $c$, $\boldsymbol{e}(c) \in [0,1]^p$, and $p$ is the number of pixels considered. $J$ denotes a piece-wise linear function, with a global minimum. It has been shown in various works such \cite{rezatofighi2019generalized,cortinhal2020salsanext}, that Lov\'asz loss is an effective additional loss term that can be used for different machine learning tasks such as object detection and segmentation. Hence, in the process of training the proposed model, Lov\'asz loss will be combined with other losses to achieve better overall accuracy of the trained model. 

Total variation (TV) regularization has been used in the literature to address the denoising problem in images \cite{chen2010adaptive, chan2005recent}. This technique forms a significant preliminary step in many computer vision tasks, such as object detection and object recognition (i.e. object localization and classification). A major concern in designing image denoising models is to preserve important image features, such as edges, while removing noise from a digital image.

The same problem can be recognized in machine learning when one tries to identify pixel or point level classification for images or point cloud data points. Although the task is not directly denoising, classifying a pixel in an image or a data point in a point cloud relies heavily on the information provided by the neighboring pixels or data points. Hence, designing a loss function based on TV regularization seems viable. In \cite{rudin1992nonlinear}, a total variation regularizer was introduced, which regularizes the overall loss by the smoothness of the prediction over neighboring pixels or data points, resulting in a better optimization for any machine learning task such as image or point cloud semantic segmentation.

Although the technique introduced in \cite{rudin1992nonlinear} and \cite{mahendran2015understanding} is mathematically sound and effective, it is only a regularization term. In this implementation, the TV regularizer is modified to a new loss function. Moreover, in combination with a standard CNN for semantic segmentation its effectiveness is shown for the task of LiDAR semantic segmentation.

Given an image-like prediction and label for the ground truth, one can write the TV loss as follows:

\begin{align}\label{eqn:einstein}
\begin{split}
    L_{tv}(y,\hat{y}) = {}& \sum_{\Delta i,\Delta j}\lVert  Y_{(\Delta i),(j)} - \hat{Y}_{(\Delta i),(j)} \rVert_{p,q} + \lVert  Y_{(i),(\Delta j)} - \hat{Y}_{(i),(\Delta j)} \rVert_{p,q} \> , \\  
    & \forall i,j, \Delta i, \Delta j \in \mathbb{Z}, p,q \geq 1
\end{split}
\end{align}

\noindent where $i,j$ are indexes for the pixel location, $\Delta i$ and $\Delta j$ are the step sizes in row and column directions, respectively. The ground truth label is denoted by $y$  and $\hat{y}$ is the prediction of the network. 
In this expression, the quantities  $Y_{(\Delta i),(j)}$ and $Y_{(i),(\Delta j)}$ represent the difference in pixel values of the current pixel with its vertical and horizontal neighbours, respectively. They are computed as,

\vspace{-1mm}
\begin{align}
    Y_{(\Delta i),(j)} = \lvert y_{(i+\Delta i), (j)} - y_{i, j}\rvert, &&
    Y_{(i),(\Delta j)} = \lvert y_{(i), (j+\Delta j)} - y_{i, j}\rvert, &&  \forall i,j, \Delta i, \Delta j \in \mathbb{Z}
\end{align}

For the task of LiDAR semantic segmentation on range-image,  equation \ref{eqn:einstein} can be simplified to the following for $L_{1,1}$ norm and one sided neighbours, i.e., $\Delta i,\Delta j=1, p,q=1$: 

\begin{align}
L_{tv}(y,\hat{y})=\sum_{i,j}| |y_{i+1,j} -y_{i,j}| - |\hat{y}_{i+1,j} -\hat{y}_{i,j}|| + ||y_{i,j+1} - y_{i,j}| - |\hat{y}_{i,j+1} - \hat{y}_{i,j}| |
\end{align}

The total loss that is used to train the proposed model is a combination of Lov\'asz, WCE, and TV losses introduced above. In order to balance the effect of each loss in the training, different weights are introduced for each loss term that can be accomodated as hyperparameters. The final loss can be formulated as
\begin{align}
    {L}_{total}= \beta_{ls}{L}_{ls} + \beta_{wce}{L}_{wce} + \beta_{tv}{L}_{tv}
\end{align}
where $\beta_{ls}$, $\beta_{wce}$ and $\beta_{tv}$ are weights for Lov\'asz loss, WCE loss, and TV loss, respectively.

	

\subsection{Training Details}

 
Before feeding the raw pointclouds $P$ into the model, they are truncated to the range [-80, 80], [-80, 80], [-5,5] in the $x$, $y$, and $z$ directions respectively. This is followed by a series of augmentations. We adopt an augmentation scheme similar to \cite{cortinhal2020salsanext}, where we do random point dropping, global pointcloud rotation and translation, and flipping along the $x$-axis. 
We drop up to 20\% of the points. For the rotations, we randomly sample angles in the range $[-5, 5]$, $[-5, 5]$, and $[-180, 180]$ degrees for $x$, $y$ and $z$ axes respectively. Similarly, for the translations, we sample in the range $[-5, 5]$, $[-3, 3]$, $[-1, 1]$ for $x$, $y$, and $z$ directions. Each augmentation is independently applied with a probability of 0.5.

The model was trained using the Adam optimizer with a one-cycle learning rate scheduler for 50 epochs. The maximum learning rate was set to 0.004, division factor of 10 and cosine annealing phase split of 0.3. A weight decay of $10^{-4}$ was also used. The models were trained on 4 Tesla V100 with per GPU batch sizes of 4 for the low-res and 3 for the high-res models respectively.

The voxel size of the PPL block was set to $[0.3125, 0.3125, 10]$ leading to a voxel grid of $[512 \times 512]$.
We used $C = 64$, $C_P = 7$ , $C_D = 192$ for filter sizes, and the height and width of the projected image were set to $H = 64$ , $W =2048$, except for the high-res model where $H = 128$.

The weights for the losses were set to $\beta_{ls}=1.5$, $\beta_{wce}=1.0$, and $\beta_{tv}=7.5$.

In the post-processing stage, the KNN used a kernel size of 5 for the low-res model and kernel size of 11 for the hi-res model, $K=5$, $\sigma=1$, and a cutoff of 1m for all models.

\section{Experimental Results}
\label{sec:result}

We use the SemanticKITTI dataset  \cite{DBLP:conf/iccv/BehleyGMQBSG19}  to evaluate the performance of our model and compare it with the state-of-the-art methods. It provides over $43K$ frames with dense point-wise annotations for the entire KITTI Odometry Benchmark across $22$ sequences. The dataset consists of 22 distinct semantic classes and is divided into two sets. The first set, referred to as the training and validation set, includes the sequences of ($00$ - $10$), while the second set referred to as the test set, includes the sequences of ($11$ - $21$). The validation set is sequence $08$ upon which the ablation studies are based upon. The point-wise labels for the first set are publicly available, however, the labels for the second set are not provided and are kept hidden for competition purposes.

In order to evaluate the results of the trained methods, the $\mathbf{mIoU}$ metric is used. $\mathbf{mIoU}$  is the most popular metric for evaluating semantic point cloud segmentation. It can be formalized as, 
\begin{equation}
            \label{slice}
         	mIoU = \frac{1}{n}\sum_{c=1}^{n}{\frac{TP_c}{TP_c+FP_c+FN_c}}
\end{equation}

where $TP_c$ is the number of true positive points for class $c$, $FP_c$ is the number of false positives, and $FN_c$ is the number of false negatives.

\subsection{Quantitative Analysis}

We compare the numerical experiments of our work with the existing methods in Table~\ref{bigtable}. It demonstrates the class wise IoU, Frames Per Second (FPS), and mean IoU for different approaches. We categorized the methods into two classes of point-wise and projection-based methods. In each category, the best IoU per class is selected. As shown, the proposed method achieves the state-of-the-art result, outperforming all the previous methods in mIoU and almost all the classes in its category. While the proposed method achieves high accuracy it is also faster than most point-based methods, making it applicable to the real-time systems.

\begin{table*}[h]
\Huge
\centering
\resizebox{\columnwidth}{!}{
\begin{tabular}{c|l|ccccccccccccccccccccc}
\hline 

\begin{sideways} Category  \end{sideways} 

& Method & \begin{sideways} Mean IoU \end{sideways} 
& \begin{sideways} Car \end{sideways} 
& \begin{sideways} Bicycle \end{sideways} 
& \begin{sideways} Motorcycle \end{sideways} 
& \begin{sideways} Truck \end{sideways} 
& \begin{sideways} Other-vehicle \end{sideways} 
& \begin{sideways} Person \end{sideways} 
& \begin{sideways} Bicyclist \end{sideways} 
& \begin{sideways} Motorcyclist \end{sideways} 
& \begin{sideways} Road \end{sideways} 
& \begin{sideways} Parking \end{sideways} 
& \begin{sideways} Sidewalk \end{sideways} 
& \begin{sideways} Other-ground \end{sideways} 
& \begin{sideways} Building \end{sideways} 
& \begin{sideways} Fence \end{sideways} 
& \begin{sideways} Vegetation \end{sideways} 
& \begin{sideways} Trunk \end{sideways} 
& \begin{sideways} Terrain \end{sideways} 
& \begin{sideways} Pole \end{sideways} 
& \begin{sideways} Traffic-sign \end{sideways} 
& \begin{sideways} FPS (Hz) \end{sideways}  \\
\hline
\multirow{9}{*}{\begin{sideways} Point-wise  \end{sideways} }
& PointNet \cite{qi2017pointnet} 
& $14.6$& $46.3$& $1.3$& $0.3$& $0.1$& $0.8$& $0.2$& $0.2$& $0.0$& $61.6$& $15.8$& $35.7$& $1.4$& $41.4$& $12.9$& $31.0$& $4.6$& $17.6$& $2.4$& $3.7$ & $2$ \\
& SPGraph \cite{landrieu2018large} 
& $20.0$& $68.3$& $0.9$& $4.5$& $0.9$& $0.8$& $1.0$& $6.0$& $0.0$& $49.5$& $1.7$& $24.2$& $0.3$& $68.2$& $22.5$& $59.2$& $27.2$& $17.0$& $18.3$& $10.5$ & $0.2$ \\
& PointNet++ \cite{qi2017pointnet++} 
& $20.1$& $53.7$& $1.9$& $0.2$& $0.9$& $0.2$& $0.9$& $1.0$& $0.0$& $72.0$& $18.7$& $41.8$& $5.6$& $62.3$& $16.9$& $46.5$& $13.8$& $30.0$& $6.0$& $8.9$ & $0.1$ \\
& SPLATNet\cite{Su_2018_CVPR} 
& $22.8$& $66.6$& $0.0$& $0.0$& $0.0$& $0.0$& $0.0$& $0.0$& $0.0$& $70.4$& $0.8$& $41.5$& $0.0$& $68.7$& $27.8$& $72.3$& $35.9$& $35.8$& $13.8$& $0.0$ & $1$ \\
& TangentConv  \cite{tatarchenko2018tangent}
& $35.9$& $86.8$& $1.3$& $12.7$& $11.6$& $10.2$& $17.1$& $20.2$& $0.5$& $82.9$& $15.2$& $61.7$& $9.0$& $82.8$& $44.2$& $75.5$& $42.5$& $55.5$& $30.2$& $22.2$ & $0.3$ \\
& PointASNL \cite{yan2020pointasnl} 
& $46.8$ & $87.9$ & $0$ & $25.1$ & $39.0$ & $29.2$ & $34.2$ & $57.6$ & $0$ & $87.4$ & $24.3$ & $74.3$ & $1.8$ & $83.1$ & $43.9$ & $84.1$ & $52.2$ & $\textbf{70.6}$ & $57.8$ & $36.9$ & $-$\\
& RandLa-Net \cite{hu2019randla} 
& $53.9$ & $94.2$ & $26.0$ & $25.8$ & $\textbf{40.1}$ & $38.9$ & $49.2$ & $48.2$ & $7.2$ & $90.7$ & $60.3$ & $73.7$ & $20.4$ & $86.9$ & $56.3$ & $81.4$ & $61.3$ & $66.8$ & $49.2$ & $47.7$  & $22$ \\
& S-BKI \cite{Gan_2020} 
& $51.3$ & $83.8$ & $30.6$ & $\textbf{43.0}$ & $26.0$ & $19.6$ & $8.5$ & $3.4$ & $0.0$ & $\textbf{92.6}$ & $\textbf{65.3}$ & $\textbf{77.4}$ & $30.1$ & $89.7$ & $63.7$ & $83.4$ & $64.3$ & $67.4$ & $\textbf{58.6}$ & $\textbf{67.1}$  & $-$\\
& Kpconv \cite{thomas2019kpconv} 
& $58.8$ & $\textbf{96.0}$ & $30.2$ & $42.5$ & $33.4$ & $\textbf{44.3}$ & $\textbf{61.5}$ & $\textbf{61.6}$ & $11.8$ & $88.8$ & $61.3$ & $72.7$ & $\textbf{31.6}$ & $\textbf{90.5}$ & $\textbf{64.2}$ & $\textbf{84.8}$ & $\textbf{69.2}$ & $69.1$ & $56.4$ & $47.4$ & $-$ \\
\hline
\multirow{13}{*}{\begin{sideways} Projection-based  \end{sideways} }
& SqueezeSeg \cite{wu2018squeezeseg} 
& $29.5$ & $68.8$ & $16.0$ & $4.1$ & $3.3$ & $3.6$ & $12.9$ & $13.1$ & $0.9$ & $85.4$ & $26.9$ & $54.3$ & $4.5$ & $57.4$ & $29.0$ & $60.0$ & $24.3$ & $53.7$ & $17.5$ & $24.5$  & $66$\\
& SqueezeSeg-CRF \cite{wu2018squeezeseg} 
& $30.8$ & $68.3$ & $18.1$ & $5.1$ & $4.1$ & $4.8$ & $16.5$ & $17.3$ & $1.2$ & $84.9$ & $28.4$ & $54.7$ & $4.6$ & $61.5$ & $29.2$ & $59.6$ & $25.5$ & $54.7$ & $11.2$ & $36.3$  & $55$ \\
& DeepTemporalSeg \cite{dewan2019deeptemporalseg} 
& $37.6$ & $81.5$ & $29.4$ & $19.6$ & $6.6$ & $6.5$ & $23.7$ & $20.1$ & $2.4$ & $85.8$ & $8.7$ & $59.3$ & $1.0$ & $78.6$ & $39.6$ & $77.1$ & $46.0$ & $58.1$ & $32.6$ & $39.1$ & $-$ \\
& SqueezeSegV2-CRF  \cite{wu2019squeezesegv2} 
& $39.6$ & $82.7$ & $21.0$ & $22.6$ & $14.5$ & $15.9$ & $20.2$ & $24.3$ & $2.9$ & $88.5$ & $42.4$ & $65.5$ & $18.7$ & $73.8$ & $41.0$ & $68.5$ & $36.9$ & $58.9$ & $12.9$ & $41.0$  & $40$\\
& SqueezeSegV2 \cite{wu2019squeezesegv2} 
& $39.7$ & $81.8$ & $18.5$ & $17.9$ & $13.4$ & $14.0$ & $20.1$ & $25.1$ & $3.9$ & $88.6$ & $45.8$ & $67.6$ & $17.7$ & $73.7$ & $41.1$ & $71.8$ & $35.8$ & $60.2$ & $20.2$ & $36.3$ & $50$\\
& SalsaNet \cite{aksoy2019salsanet} 
& $45.4$ & $87.5$ & $26.2$ & $24.6$ & $24.0$ & $17.5$ & $33.2$ & $31.1$ & $8.4$ & $89.7$ & $51.7$ & $70.7$ & $19.7$ & $82.8$ & $48.0$ & $73.0$ & $40.0$ & $61.7$ & $31.3$ & $41.9$ & $26$ \\
& RangeNet21 \cite{milioto2019rangenet++} 
&  $47.4$ & $85.4$ & $26.2$ & $26.5$ & $18.6$ & $15.6$ & $31.8$ & $33.6$ & $4.0$ & $91.4$ & $57.0$ & $74.0$ & $26.4$ & $81.9$ & $52.3$ & $77.6$ & $48.4$ & $63.6$ & $36.0$ & $50.0$  & $20$ \\
& RangeNet53 \cite{milioto2019rangenet++} 
&  $49.9$ & $86.4$ & $24.5$ & $32.7$ & $25.5$ & $22.6$ & $36.2$ & $33.6$ & $4.7$ & $91.8$ & $64.8$ & $74.6$ & $27.9$ & $84.1$ & $55.0$ & $78.3$ & $50.1$ & $64.0$ & $38.9$ & $52.2$
 & $13$ \\
& LatticeNet  \cite{rosu2019latticenet} 
& $52.9$ & $92.9$ & $16.6$ & $22.2$ & $26.6$ & $21.4$ & $35.6$ & $43.0$ & $\textbf{46.0}$ & $90.0$ & $59.4$ & $74.1$ & $22.0$ & $88.2$ & $58.8$ & $81.7$ & $63.6$ & $63.1$ & $51.9$ & $48.4$  & $7$\\
& RangeNet53++KNN \cite{milioto2019rangenet++} 
& $52.2$ & $91.4$ & $25.7$ & $34.4$ & $25.7$ & $23.0$ & $38.3$ & $38.8$ & $4.8$ & $91.8$ & $65.0$ & $75.2$ & $27.8$ & $87.4$ & $58.6$ & $80.5$ & $55.1$ & $64.6$ & $47.9$ & $55.9$ & $12$  \\
& PolarNet \cite{zhang2020polarnet} 
& $54.3$ & $93.8$ & $40.3$ & $30.1$ & $22.9$ & $28.5$ & $43.2$ & $40.2$ & $5.6$ & $90.8$ & $61.7$ & $74.4$ & $21.7$ & $90.0$ & $61.3$ & $84.0$ & $65.5$ & $67.8$ & $51.8$ & $57.5$ & $16$ \\
& 3D-MiniNet-KNN \cite{alonso20203d} 
& $55.8$ & $90.5$ & $42.3$ & $42.1$ & $28.5$ & $29.4$ & $47.8$ & $44.1$ & $14.5$ & $91.6$ & $64.2$ & $74.5$ & $25.4$ & $89.4$ & $60.8$ & $82.8$ & $60.8$ & $66.7$ & $48.0$ & $56.6$  & $28$\\
& SqueezeSegV3-53 \cite{xu2020squeezesegv3} 
& $55.9$ & $92.5$ & $38.7$ & $36.5$ & $29.6$ & $33.0$ & $45.6$ & $46.2$ & $20.1$ & $91.7$ & $63.4$ & $74.8$ & $26.4$ & $89.0$ & $59.4$ & $82.0$ & $58.7$ & $65.4$ & $49.6$ & $58.9$ & $6$  \\
& SalsaNext \cite{cortinhal2020salsanext} 
& $59.5$ & $91.9$ & $\textbf{48.3}$ & $38.6$ & $38.9$ & $31.9$ & $60.2$ & $59.0$ & $19.4$ & $91.7$ & $63.7$ & $75.8$ & $29.1$ & $90.2$ & $\textbf{64.2}$ & $81.8$ & $63.6$ & $66.5$ & $54.3$ & $62.1$  & $24$  \\

\cline{2-23} 
& \textbf{TORNADONet} [Ours] 
&$\textbf{61.1}$ & $\textbf{93.1}$ & $\textbf{53.0}$ & $\textbf{44.4}$ & $\textbf{43.1}$ & $\textbf{39.4}$ & $\textbf{61.6}$ & $56.7$ & $20.2$ & $90.8$ & $\textbf{65.3}$ & $75.3$ & $27.5$ & $89.6$ & $62.9$ & $\textbf{84.1}$ & $\textbf{64.3}$ & $\textbf{69.6}$ & $\textbf{55.0}$ & $\textbf{64.2}$ & $7$ \\

& \textbf{TORNADONet-HiRes} [Ours] 
&$\textbf{63.1}$ & $\textbf{94.2}$ & $\textbf{55.7}$ & $\textbf{48.1}$ & $\textbf{40.0}$ & $\textbf{38.2}$ & $\textbf{63.6}$ & $\textbf{60.1}$ & $34.9$ & $89.7$ & $\textbf{66.3}$ & $74.5$ & $28.7$ & $\textbf{91.3}$ & $\textbf{65.6}$ & $\textbf{85.6}$ & $\textbf{67.0}$ & $\textbf{71.5}$ & $\textbf{58.0}$ & $\textbf{65.9}$ & $4$  \\

\hline
\end{tabular}
}
\caption[CPNET]{IoU results on the Semantic-Kitti dataset test split. FPS measurements were taken using a single GTX 2080Ti GPU, or approximated if a runtime comparison was made on another GPU. Note that a FPS of $10$ or more is considered real-time, since the acquisition frequency of the Velodyne HDL-64E 64 beam LiDAR sensor is 10 Hz.}
\label{bigtable}
\end{table*}

\subsection{Qualitative Analysis}
\begin{figure} [!htbp]
	\centering
	\begin{subfigure}[t]{0.38\textwidth}
		\includegraphics[width=\textwidth]{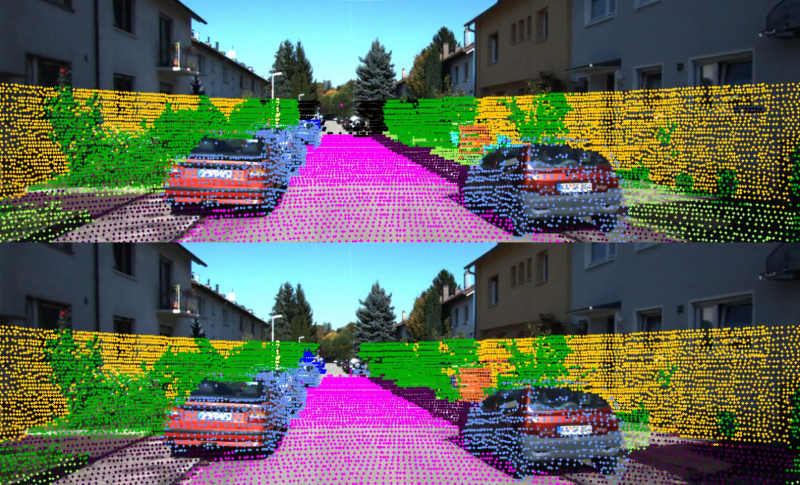}
		\caption{Case 1}
		\label{fig:case 1}
	\end{subfigure}
	~ 
	\begin{subfigure}[t]{0.38\textwidth}
		\includegraphics[width=\textwidth]{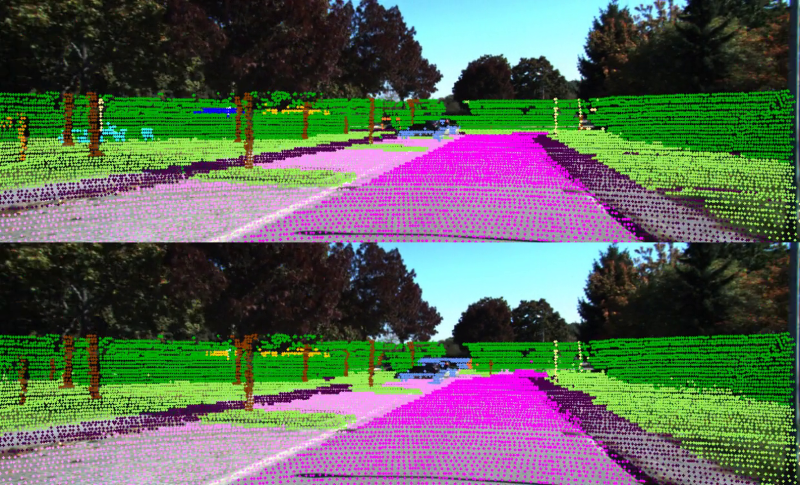}
		\caption{Case 2}
		\label{fig:case 2}
	\end{subfigure}
	~ 
	\begin{subfigure}[t]{0.38\textwidth}
		\includegraphics[width=\textwidth]{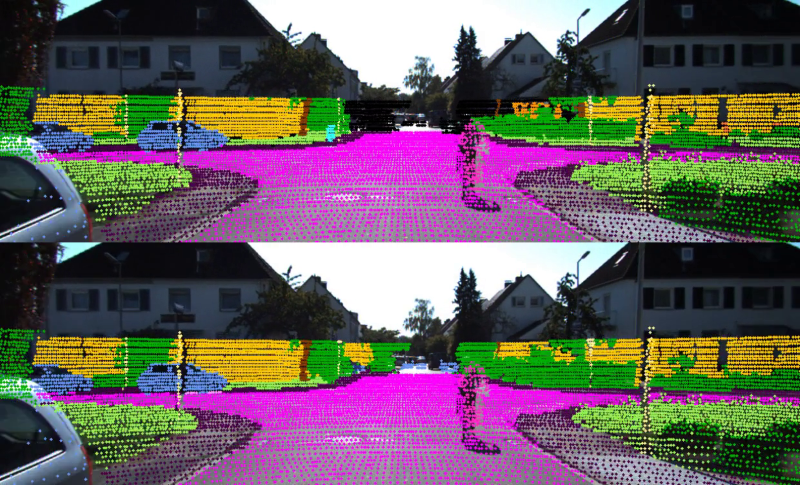}
		\caption{Case 3}
		\label{fig:case 3}
	\end{subfigure}
	~
	\begin{subfigure}[t]{0.38\textwidth}
		\includegraphics[width=\textwidth]{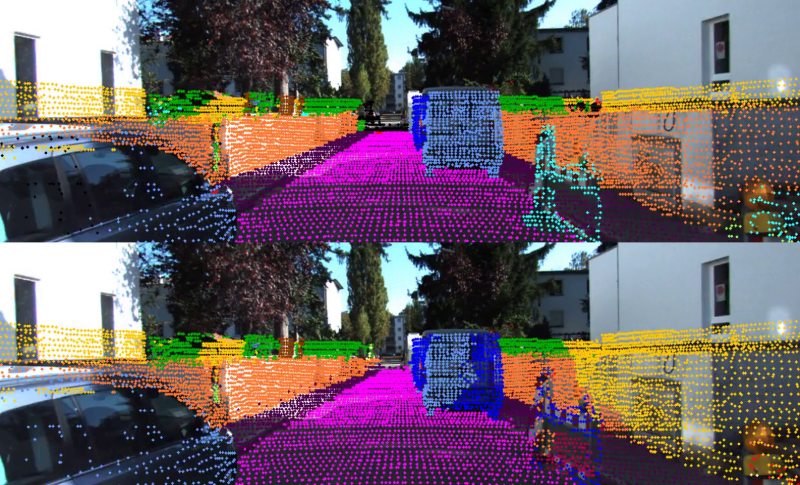}
		\caption{Case 4}
		\label{fig:case 4}
	\end{subfigure}
	\caption{Ground truth (top) and prediction (bottom) on SemanticKITTI validation set projected onto the camera images. \ref{fig:case 1}, \ref{fig:case 2}, \ref{fig:case 3} show successful semantic predictions, while \ref{fig:case 4} highlights the challenge of distinguishing between 2 similar classes.}
	\label{fig:Qualitative}
\end{figure}

In order to better understand the results produced by the proposed method, TORNADO-Net, in this section, four different samples from the SemanticKITTI \cite{DBLP:conf/iccv/BehleyGMQBSG19} validation set are provided in Figure~\ref{fig:Qualitative}. The results are presented along with ground truth labels for each case presenting the quality of the results in different situations where different objects/scene are available. Case $1$ shows how well TORNADO-Net can perform where there exist many dynamic objects such as cars (parked or otherwise). Cases $2$ and $3$ depicts the successful prediction for road, vegetation, pole and other structural objects. As it is shown, TORNADO-Net can handle most of the cases well and in many cases, distinguishing between ground truth and prediction is hard.

However, there are situations for which TORNADO-Net fails to predict the correct class in a scene. Case $4$ presents one of the more common failures seen in our analysis. In this case, the truck on the middle-right part of the scene is predicted partially as a car and partially as a truck. This is mostly due to limited number of training samples and a fair amount of shared features between different types of vehicles such as overall structure, intensity reflection, etc. Nevertheless, the authors think these issues should be addressed even with limited labeled data and is part of the future research on this topic. For more thorough analysis of the results a video of the predictions on sequence $8$ will be available publicly as supplementary material.
 

\subsection{Ablations}
    As shown in Table \ref{tab:Ablation}, each of our contributions amounts to an improvement in the overall mIoU. We receive the biggest increases with the introduction of the TV Loss. Circular padding mostly had a positive effect (about 0.5\%). The PPL block is also a key component introducing a 1.5\% jump at the cost of extra parameters and reduction in speed. Using a hi-res version of the same model leads to a further improvement in the mIoU at the cost of speed. KNN post-processing is also a key component leading to a 2-3\% jump. The high-res model benefits from a higher KNN search value due to the sparser projection of points onto the image.

\begin{table}[h]
\tabcolsep=0.11cm
\begin{center}
\scalebox{0.7}{
{\tabulinesep=0.8mm 
\begin{tabu}{c|cccccc|ccc}
\hline \hline 
\multicolumn{1}{c|}{\textbf{ Architecture }} & \multicolumn{1}{c}{\textbf{\begin{turn}{30} Augm.\end{turn} }} & \multicolumn{1}{c}{\textbf{\begin{turn}{30} Diamond \end{turn}}} &
\multicolumn{1}{c}{\textbf{\begin{turn}{30} TV Loss \end{turn}}} &
\multicolumn{1}{c}{\textbf{\begin{turn}{30} Circ. Padd. \end{turn}}} &
\multicolumn{1}{c}{\textbf{\begin{turn}{30} PPL \end{turn}}} &
\multicolumn{1}{c|}{\textbf{\begin{turn}{30} Hi-Res \end{turn}}} &

\multicolumn{1}{c}{\textbf{\begin{tabular}[c]{@{}c@{}}mIoU\\ No KNN\end{tabular}}} & \multicolumn{1}{c}{\textbf{\begin{tabular}[c]{@{}c@{}}mIoU\\ KNN 5\end{tabular}}} & 
\multicolumn{1}{c}{\textbf{\begin{tabular}[c]{@{}c@{}}mIoU\\ KNN 11\end{tabular}}}\\
 \hline \hline 
\multirow{2}{*}{Baseline}
&  &  &   &   & & & 55.0 & 58.1 & 58.1\\ 
&  \checkmark  & &  &   & &  & 55.1 & 58.3 & 58.4\\ \hline
\multirow{7}{*}{TORNADONet}
& \checkmark  & \checkmark & &  & &  & 55.1 & 58.3 & 58.4\\
& \checkmark  & \checkmark & \checkmark &  & &  & 57.3 & 60.9 & 60.8\\
& \checkmark  & \checkmark & \checkmark & \checkmark & &  & 57.8 & 61.3 & 61.2\\
& \checkmark  & \checkmark  & \checkmark &  & \checkmark & & 58.8 & 62.5 & 62.5\\ 
& \checkmark  & \checkmark  & \checkmark & \checkmark & \checkmark & & 58.3 & 62.0 & 61.9\\ 
& \checkmark  & \checkmark & \checkmark &  & \checkmark & \multicolumn{1}{c|}{\checkmark}  & 61.2 & 63.6 & 63.9 \\
& \checkmark  & \checkmark & \checkmark & \checkmark & \checkmark & \multicolumn{1}{c|}{\checkmark}  & 61.8 & 64.2 & 64.5 \\ \hline
\end{tabu}}}
\end{center}
\caption{Ablative Analysis evaluated on SemanticKITTI dataset validation (seq 08). }
\label{tab:Ablation}
\end{table}

\section{Conclusion}
\label{sec:conclusion}
	
Semantic segmentation has been a subject of interest in many fields, such as autonomous driving. While other deep learning techniques are promising on the LiDAR semantic segmentation task, they are complex systems to implement. However, other methods that are real-time systems lack accurate performance. The main objective of this work was to propose a novel deep neural network, TORNADO-Net, for 3D LiDAR point cloud semantic segmentation.  We leveraged a multi-view (bird-eye and range) projection feature extraction with an encoder-decoder ResNet architecture with a novel diamond context block. Moreover, TV loss was introduced along with Lov\'asz-Softmax, and WCE loss to efficiently train the network. We evaluated the proposed method on the SemanticKITTI benchmark and were able to achieve state-of-the-art results on its published leaderboard, outperforming all the previous methods. 



\clearpage



{\small
\bibliographystyle{unsrt}
\bibliography{tornado_net}
}
\end{document}